%% file: mwextractor.tex
\documentclass[11pt]{article}

\usepackage{emnlp2022}

\usepackage{times}
\usepackage{latexsym}
\usepackage{amssymb}

\usepackage[T1]{fontenc}

\usepackage[utf8]{inputenc}

\usepackage{microtype}
\usepackage{multirow}
\usepackage{booktabs,arydshln}

\usepackage{inconsolata}

\makeatletter
\def\adl@drawiv#1#2#3{%
        \hskip.5\tabcolsep
        \xleaders#3{#2.5\@tempdimb #1{1}#2.5\@tempdimb}%
                #2\z@ plus1fil minus1fil\relax
        \hskip.5\tabcolsep}
\newcommand{\cdashlinelr}[1]{%
  \noalign{\vskip\aboverulesep
           \global\let\@dashdrawstore\adl@draw
           \global\let\adl@draw\adl@drawiv}
  \cdashline{#1}
  \noalign{\global\let\adl@draw\@dashdrawstore
           \vskip\belowrulesep}}
\makeatother

\usepackage{array}
\newcolumntype{P}[1]{>{\centering\arraybackslash}p{#1}}

\input{tikz}

\title{Unsupervised Term Extraction for Highly Technical Domains}

\author{Francesco Fusco\\
IBM Research \\
\texttt{ffu@zurich.ibm.com} \\\And
Peter Staar\\
IBM Research\\
\texttt{taa@zurich.ibm.com} \\\And
Diego Antognini\\
IBM Research\\
\texttt{Diego.Antognini@ibm.com} \\}

\begin{document}
\maketitle

\begin{abstract}



Term extraction is an information extraction task at the root of
knowledge discovery platforms. Developing term extractors that are able to
generalize across very diverse and potentially highly technical
domains is challenging, as annotations for domains requiring in-depth
expertise are scarce and expensive to obtain.  In this paper, we describe
the term extraction subsystem of a commercial knowledge discovery platform
that targets highly technical fields such as pharma, medical, and material
science. To be able to generalize across domains, we introduce a \textit{fully
    unsupervised} annotator~(UA). It extracts terms by combining novel morphological signals from sub-word tokenization with term-to-topic and intra-term
similarity metrics, computed using general-domain pre-trained sentence-encoders.
The annotator is used to implement a \textit{weakly-supervised setup}, where
transformer-models are fine-tuned    (or pre-trained) over the training
data \textit{generated} by running the UA over large unlabeled corpora.
Our experiments demonstrate that our setup can improve the
predictive performance \textit{while decreasing} the inference latency on
both CPUs and GPUs. Our annotators provide a very competitive baseline for
all the cases where annotations are not available.


%


\end{abstract}

\section{Introduction}


Automated Term Extraction~(ATE) is the task of extracting terminology from
domain-specific corpora. Term extraction is the most important information
extraction task for knowledge discovery systems -- whose aim is to create structured knowledge from unstructured text -- because domain specific terms are
the linguistic representation of domain-specific concepts. To be of use in
knowledge discovery systems (e.g., SAGA~\cite{continuous-construction}, DeepSearch~\cite{dognin-etal-2020-dualtkb}) the term extraction has to identify individual
\textit{mentions} of terms to enable downstream components~(i.e., the entity
linker) to use not only the terms, but also their surrounding context. Unlike
other applications of term extraction, such as text classification,
where it is sufficient to extract representative terms
for entire documents or even use generative approaches, term extraction in
knowledge discovery systems has to be approached as a sequence tagging task.

\begin{figure}[t!]
    \centering
    \includegraphics[width=0.475\textwidth,height=4.61cm]{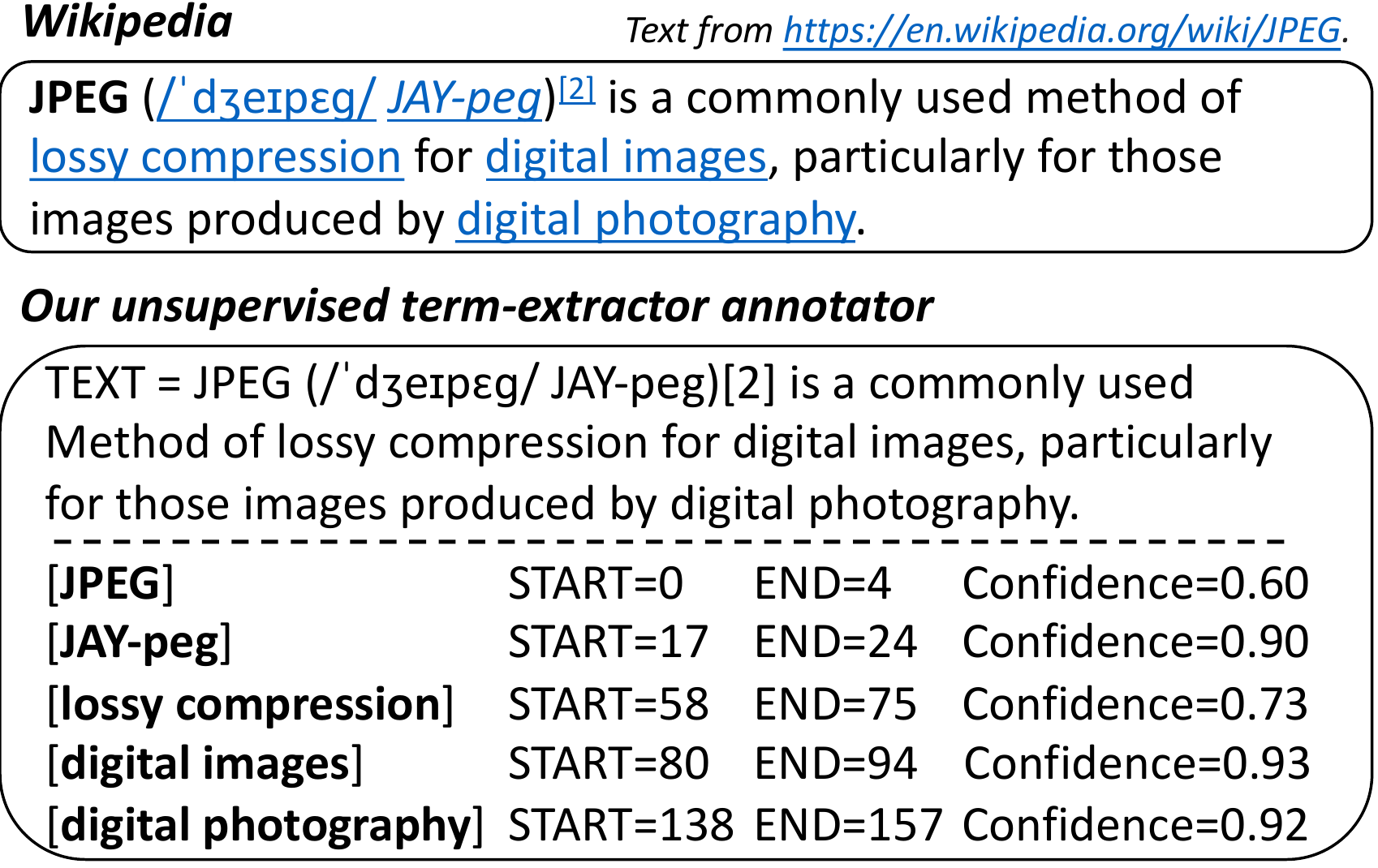}
    \caption{Our term extractor identifies the same mentions as Wikipedia without \textit{relying on annotated data}.}
    \label{fig:jpeg}
\end{figure}

The largest challenges for term extraction systems, when used for knowledge
discovery, are generalization across domains and lack of annotated data.  In
fact, commercial knowledge discovery platforms are typically required to
process large corpora targeting very diverse and often highly technical
domains. Organizing annotation campaigns for such vertical domains is a costly
process as it requires highly specialized domain experts. An additional
challenge for such platforms are the computational requirements, which must be
accounted for when developing technologies required to sift through very large
and often proprietary corpora.

In this work, we describe an effective term
extraction approach used in a commercial knowledge discovery platform\footnote{\url{https://ds4sd.github.io}.} to
extract \textit{Wikipedia-like concepts}\footnote{The linking from words to Wikilinks is done manually on Wikipedia, see \url{https://en.wikipedia.org/wiki/Wikipedia:Manual_of_Style/Linking} for more details.} from text ~(see Figure~\ref{fig:jpeg}).
Our approach does not require any human annotation, offers the flexibility to
select the right trade-off between accuracy and inference latency, and
enables the deployment of lightweight models running entirely on CPUs.



\begin{figure}[!t]
    \centering
    \includegraphics[width=0.485\textwidth]{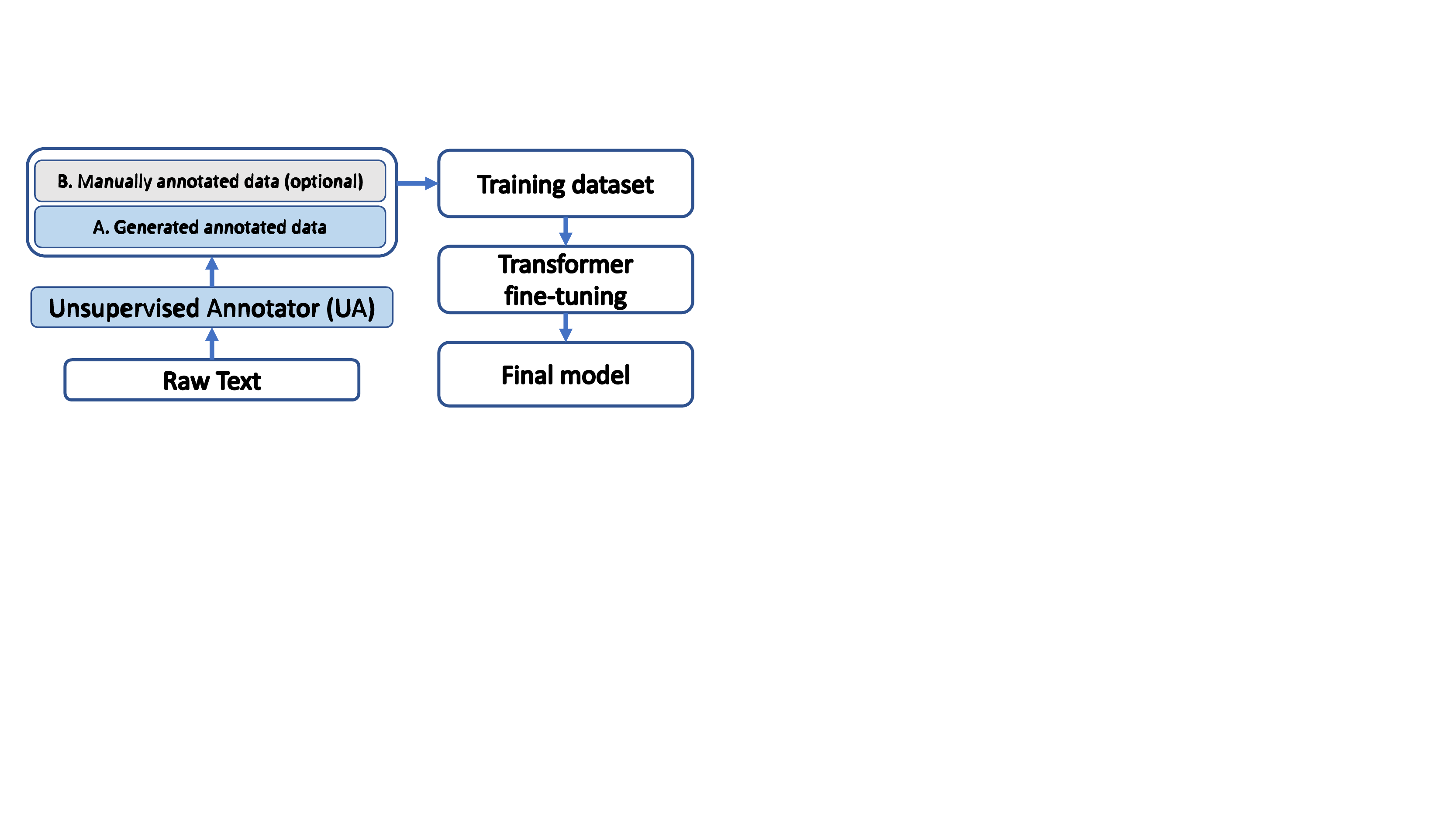}
    \caption{Our training workflow consists~of 1) generating training data from raw unlabeled text using our Unsupervised Annotator, and 2) fine-tuning a transformer-based model or any sequence tagging~model.}
    \label{fig:training-workflow}
\end{figure}

At its core, our approach is a \textit{weakly supervised} setup (see Figure~\ref{fig:training-workflow}), where
transformer models are fine-tuned~(or even entirely pre-trained) using the
\textit{weak labels} generated by a \textit{fully unsupervised} term annotator.
The unsupervised annotator~(UA) combines novel morphological and semantic
signals to tag sequences of text corresponding to domain-specific terminology.
In fact, in addition to part-of-speech tagging to identify candidate terms, the
UA exploits sub-word tokenization techniques -- commonly used in language models to highlight words that are outside of the common vocabulary -- to indirectly measure the morphological complexity of a word based on its sub-tokens. To the
best of our knowledge, this is the first work relying on sub-word tokenization
units in the context of term extraction. To prune the candidate set of terms
the annotator uses two semantic metrics as thresholds: the \textit{topic-score}
and a novel \textit{specificity score} that are both computed using representations from sentence encoders. The unsupervised annotator, combined with the
two-stage weakly supervised setup, makes our approach particularly attractive
for practical industrial setups because computationally intensive techniques used by
the unsupervised annotator are not paid at inference time. Therefore, one
can improve the annotation quality by using more expensive techniques (e.g.,
entity linking to external knowledge bases), without adding costs at inference
time. The two main contributions of this paper are
summarized as follows:
\begin{enumerate}
    \item We extract a novel morphology signal from subword-unit tokenization
          and we introduce a new metric called the \textit{specificity score}.
    Upon those signals, we build an unsupervised term-extractor that offers competitive results when no
          annotation is available.
	\item We show that by fine-tuning transformer models over the weak labels produced by the unsupervised
          term extractor we decrease the latency and improve the prediction quality.
\end{enumerate}

\section{Related work}                                                                 
                                                                                        
Automated Term Extraction~(ATE) is a natural language processing task that has         
been the subject of many research                                                           
studies~\cite{ontlearn, DBLP:journals/ir/VenturaJRT16,ZHANG2018102,ma-etal-2019-exploring,sajatovic-etal-2019-evaluating}.
What we describe in this work is an effective term extraction approach that is fully    
unsupervised and also offers the flexibility and modularity to deploy and easily    
maintain systems in production.                                                         
                                                                                        
ATE should not be confused with keyphrase                                               
extraction~\cite{firoozeh_nazarenko_alizon_daille_2020,                                 
mahata-etal-2018-key2vec, bennani-smires-etal-2018-simple} and keyphrase                
generation~\cite{DBLP:journals/corr/abs-2203-08118, chen-etal-2020-exclusive},          
which have the goal of extracting, or generating, key phrases that best                 
describe a given free text document. Keyphrases can be seen as a set of                  
tags associated to a document. In the context of keyphrase                 
extraction, sentence embedders have been used in the literature, such as in                 
EmbedRank~\cite{bennani-smires-etal-2018-simple} and                                    
Key2Vec~\cite{mahata-etal-2018-key2vec}. In our work, we also rely on sentence          
encoders, but we use them to generate training data~for~sequence tagging.          
Therefore, we do not rely on sentence encoders at runtime to extract               
terminology from text, enabling the creation of lower latency systems.                                                                                
                                                                                        
To capture complex morphological structures~we use word segmentation                                                    
techniques.  Word segmentation algorithms such as Byte-Pair                             
Encoding \cite{sennrich2016}, word-piece~\cite{37842}, and unigram language              
modeling \cite{kudo-2018-subword} have been introduced to avoid the problem of          
out-of-vocabulary words and, more in general, to reduce the number of distinct          
symbols that sequence models for natural language processing have to process.           
To the best of our knowledge, we are the first to use the subword-unit tokenization as a signal to extract technical terms from text.                    
                                                                                        
Our approach builds on the notion of specificity to find terminology. While           
there are multiple research                                                             
works~\cite{caraballo-charniak-1999-determining,ryu-choi-2006-taxonomy}                 
highlighting the importance of specificity, to the best of our knowledge, this           
is the first work using the notion of specificity to extract terminology from           
text.

\section{The approach}

Figure~\ref{fig:training-workflow} depicts our \textit{weakly supervised}
setup.  Starting from a raw text corpus and no labels, our training workflow
produces an efficient sequence tagging model, based on the transformer architecture, which
effectively implements the term extraction. At the core of the \textit{weak
labels} there is a fully unsupervised component, called the
\textit{Unsupervised Annotator}~\textit{(UA)}, which, given the raw corpus, produces a training dataset for sequence labeling. The resulting dataset is
used to train~(or fine-tune) a sequence model that represents the final model
for term annotation used at inference time. Pre-trained transformer-based
models clearly represent a valid alternative to implement such sequence models. Moreover, we can avoid pre-training since the UA potentially generates a large amount of training data.

From the software engineering standpoint, this setup is extremely attractive as
it makes the architecture of the term extraction subsystem modular and very
flexible. The modularity comes from decoupling
the inference component and the unsupervised annotator~(UA). The
unsupervised annotator can be enhanced with additional and more computationally
demanding subcomponents~(e.g., an entity linker to an external knowledge base),
without increasing the final inference latency observed by the user. This
modularity enables domain customization with proprietary data~(and systems), 
which might be available for specific domains or customers. Since the
integration between the Unsupervised Annotator and the inferencing component is
achieved via data (i.e., the training samples for sequence tagging expressed in IOB
format) the approach enables the smooth transition between a fully unsupervised
setup and a setup where manual annotations augment the ones obtained via the UA.
In practice, in realistic deployments, the unsupervised annotator is used to
\textit{boostrap} the term extraction subsystem, while domain specific
annotations are added over time by organizing annotation campaigns or by
collecting labels through the interactions of the users with the knowledge
discovery platform.

Having a dedicated component for inferencing, which is independent from the UA,
gives the flexibility to select the right trade-off in terms of accuracy,
inference latency, deployment costs, and inferencing infrastructure. This
choice is completely independent from the Unsupervised Annotator, which can be
independently improved without taking care of inference latency.
Since the inference component can be built around off-the-shelf
transformer-based models, one can fully leverage the optimizations available in
modern commercial offerings for inferencing services~(e.g., Amazon Sagemaker, HuggingFace Infinity).  As
Transformer-based models are frequently used for multiple tasks~(e.g., classification,
NER, QA) within a knowledge discovery platform, this often corresponds to
having a very homogeneous inferencing infrastructure in production.  However,
given that the UA can potentially generate a large amount of training samples,
large pre-trained models are not a necessity, and even alternative
architectures such as pQRNN~\citep{DBLP:journals/corr/abs-2101-08890} or pNLP-Mixer~\citep{fusco2022pnlp} can be used.

\subsection{Unsupervised annotator}


Our unsupervised annotator is responsible for providing accuracy in potentially
unseen domains \textit{without} any training data, as depicted in
Figure~\ref{fig:jpeg}. It achieves this goal by using a greedy
approach that processes each sentence of a raw corpus using the following steps:



\smallskip
\noindent
\textbf{1. Extract multiword expression candidates}.~Using the part-of-speech tags we extract~multiword~expression candidates, consisting of
sequences of zero or more adjectives (ADJ) followed by nouns (NOUN) or proper
nouns (PROPNs) sequences. This chunking step allows us to identify
term candidates expressed via multiword expressions.

\smallskip
\noindent
\textbf{2. Filter candidates by specificity or topic score}.  Once the candidate
terms, represented as multiword expression, are identified, a pruning step is responsible
for filtering out multiword expressions using two semantic scores: the \textit{topic score}
and the \textit{specificity score}. To compute those scores, we rely on pre-trained
sentence encoders to extract embeddings from text.

\noindent
$\blacksquare$ \textbf{Topic score.} The topic score captures the similarity,
topic-wise, between a candidate and the sentence containing it. It is computed
as the cosine similarity between the embedding vector of the multiword
expression and the embedding vector of the sentence containing
it.

\noindent
$\blacksquare$ \textbf{Specificity score ($SP$).} This is the mean of the
pairwise distance, in the embedding space, between the multiword expressions
and all the other word or multiword expression in the context. Specifically, given a multiword ${mw}$, and the word or multiword expression $w_1,
... , w_k$ in its context, we define the specificity score $SP$ as:\begin{equation}
SP({mw}) = {{ \sum_{i=1}^{k} {dist( {w_i}, {mw} ) } } \over {  k }},	
\end{equation}where $dist({w_i}, {w_j})$ is the cosine-similarity between the embedding
vectors of ${w_i}$ and
${w_j}$. Multiword expressions with a higher score correspond to more specific terms.

\textit{Multiword expressions with a specificity or topic score below a certain threshold
can be filtered out.} Both scores rely on high-quality sentence encoders.  In
our implementation we use the pretrained sentence encoders described in
\citet{reimers-2019-sentence-bert}, but other sentence encoders can be used as a
drop-in replacement.


\smallskip
\noindent
\textbf{3. Upgrade single nouns according to morphological features}. At this
stage, we could have nouns that are not part of any multiword expressions, but
still relevant. We deal with those cases separately. For each of those nouns,
we have to decide whether to extract them as terms or not. To do so, we use
morphological features. First, we check if the lemma of the noun is the same as
any of the heads of the multiword expressions. If that is the case, we upgrade
the noun to term. Otherwise, we segment the word using a subword-unit
segmentation algorithm and a vocabulary trained over a large general purpose
corpus.  Subword-unit tokenizers have been introduced to enable the representation of any text as a combination of subword units, with the idea that the most frequent
words can be represented by a small number of subword units, eventually just
one for very common words as in case for stopwords. For example, the word
\textit{``sun''}, will have its own entry in the dictionary of subword units,
while the word \textit{``paracetamol''} will be represented as the sequence of
the following subword units:[ \textit{``para'', ``\#\#ce'', ``\#\#tam'', ``\#\#ol''}].
Not suprisingly, the number of subword units required to represent a word in a
subword-unit tokenization regime is a very strong morphological signal, which
we use as an \textit{indirect measure} of the morphological ``complexity'', and is extremely cheap to compute. In our implementation, we
simply promote as terms all the nouns with a number of sub-tokens higher than
a threshold~(4 in our case). We use the vocabulary of the
BERT-base model from HuggingFace~\cite{wolf-etal-2020-transformers} and the
corresponding~tokenizer.


\section{Experiments}
\begin{table}[t!]
	\small
    \centering
    \begin{tabular}{@{}l@{}
    r@{\hspace{1.5mm}}r@{\hspace{1.5mm}}r@{}c@{\hspace{4mm}}
    r@{\hspace{1.5mm}}r@{\hspace{1.5mm}}r@{}}
        & \multicolumn{3}{c}{Sentence} & & \multicolumn{3}{c}{Terms} \\
        \cmidrule{2-4}\cmidrule{6-8}
        Corpus & Train & Dev & Test & & Train & Dev & Test \\
     	\toprule  
        ACL & $828$ & $276$ & $280$ & & $2,574$ & $898$ & $930$ \\
        GENIA & $11,127$ &  $3,709$ & $3,710$ & & $48,928$ & $16,217$ & $16,404$\\
        ScienceIE & $2,516$ & $417$ & $876$ & & $6,067$ & $1,052$ & $1,885$\\
    \end{tabular}
    \caption{Number of sentences and terms in the train, dev, and test set for the datasetst used for evaluation.}
    \label{tab:datasummary}
\end{table}
\begin{table*}
\small
\centering
    \begin{tabular}{
    @{}l@{}
    c@{\hspace{0mm}}
    c@{\hspace{1mm}}c
    @{}}
         & & \multicolumn{2}{c}{ACL}\\
        \cmidrule{3-4}
        Model (\#Params)& &  exact $F_1$ & partial $F_1$\\
\toprule
BERT B (110M) & & $78.69$ & $91.06$ \\
ELECTRA S (14M) &  & $72.84$ & $88.06$\\
ELECTRA XS (7M) & & $50.40$ & $71.61$\\
\cdashlinelr{1-4}
UA (0) & & $49.95$ & $74.56$ \\
    \end{tabular}\hfill
        \begin{tabular}{
    @{}l@{}
    c@{\hspace{0mm}}
    c@{\hspace{1mm}}c
    @{}}
         & & \multicolumn{2}{c}{GENIA}\\
        \cmidrule{3-4}
        Model (\#Params)& &  exact $F_1$ & partial $F_1$\\
\toprule
BERT B (110M) & & $70.13$ & $88.19$ \\
ELECTRA S (14M) &  & $67.73$ & $88.04$\\
ELECTRA XS (7M) & & $59.86$ & $83.16$\\
\cdashlinelr{1-4}
UA (0) & & $45.65$ & $77.16$  \\
    \end{tabular}\hfill
        \begin{tabular}{
    @{}l@{}
    c@{\hspace{0mm}}
    c@{\hspace{1mm}}c
    @{}}
         & & \multicolumn{2}{c}{ScienceIE}\\
        \cmidrule{3-4}
        Model (\#Params)& &  exact $F_1$ & partial $F_1$\\
\toprule
BERT B (110M) & & $49.62$ & $66.36$ \\
ELECTRA S (14M) &  & $46.43$ & $68.45$\\
ELECTRA XS (7M) & & $27.17$ & $51.10$\\
\cdashlinelr{1-4}
UA (0) & & $39.75$ & $64.29$ \\
    \end{tabular}
    \caption{\label{tab:result_summary_train}Results for the unsupervised annotator~(UA) and transformer models fine-tuned on the \textbf{manually annotated} ACL, GENIA, and ScienceIE datasets, respectively. Without using any annotation, the UA performs similarly to ELECTRA XSmall and even better on the ScienceIE.}
\end{table*}

We now assess whether our approach can represent
a \textit{valid baseline} for term extraction in different technical domains
when annotated data is \textit{not available}. We aim to answer the following research~questions: \begin{itemize}
 	 \item Does our Unsupervised Annotator generate a high-quality weakly-annotated dataset from a unlabeled general-domain corpus?
 	 \item Can we train models on the latter to lower the latency inference \textbf{and} increase the prediction performance at the same time?
 \end{itemize}

\subsection{Datasets}

We use three common publicly available term extraction corpora:
ACL RD-TEC 2.0 \cite{corpus:acl}, GENIA \cite{corpus:genia}, and ScienceIE \cite{corpus:scienceie}. Each contains abstracts from scientific articles in different domains: natural language processing (ACL), medicine (GENIA), and computer science, material science, as well as physics (ScienceIE). All tokens are annotated using the IOB format (short for Inside, Out and Begin)~\cite{ramshaw1999text}. Since we are only interested in general term extraction, we did not use multiple class
labels, even if provided in the respective dataset. We create random splits of train, dev, and test sets~(60/20/20) for the ACL and GENIA datasets, and we use the pre-existing data splits for ScienceIE corpus. 

In terms of preprocessing, we remove nested terms from the GENIA dataset, since the IOB tag set does not allow nested term extraction. For the ACL corpus, some samples have abstracts labeled by two annotators. In those cases, we selected the abstract from the first annotator. An overview of the datasets is given in Table~\ref{tab:datasummary}.

Since our objective is to study the generalization of our approach, we need an \textit{unlabeled} broad corpus from which our Unsupervised Annotator will annotate the text. Hence, we randomly sampled 500,000 sentences from abstracts from Semantic Scholar~(SS). \footnote{\href{https://www.semanticscholar.org/}{www.semanticscholar.org/}.} We call our weakly annotated training set UA-SS. The training sets of the ACL, GENIA, and ScienceIE datasets are not used (unless specified).

\subsection{Models}

We use transformer-models, fine-tuned with manual annotations, as baselines. We employ pre-trained transformer models of different sizes: BERT-base (110M parameters)~\cite{devlin-etal-2019-bert}, ELECTRA Small (14M parameters)~\cite{clark2020electric}, and ELECTRA XSmall (7M parameters). 

Since our main goal is to compare the models to each other and across multiple corpora, we prioritize comparabability across corpora over comparabability with approaches from other studies.

\subsection{Experimental settings}

We use the pre-trained checkpoints of BERT-base 
and ELECTRA Small 
from HuggingFace~\cite{wolf-etal-2020-transformers}. We pre-train ELECTRA XSmall\footnote{We used 2 attention heads and 4 hidden layers, while using the same hidden dimension and similarly sized vocabulary.} from scratch using our Semantic Scholar dataset. 
During fine-tuning, we devoted a similar amount of
GPU time to all the models. We pick the best-performing model in the
dev set after 10 epochs

We implemented our Unsupervised Annotator using the POS tagger of SpaCy~\cite{spacy}. 
To compute the specificity and similarity scores we use the sentence embedding model \texttt{distilbert-base-nli-mean-tokens} from the
\texttt{sentence transformers}\footnote{\href{https://pypi.org/project/sentence-transformers/}{pypi.org/project/sentence-transformers/}.}
library.

The specificity and similarity thresholds used to generate the training data
over abstracts from Semantic Schoolar have been set to conservative values. We set the threshold for the specificity $T_{SP}=0.05$
and the threshold for the similarity $T_{topic}=0.1$.
For the sub-word tokenization we rely on the tokenizer from BERT-base.
 
\begin{table*}
\small
\centering

    \begin{tabular}{
    @{}P{2.5cm}@{\hspace{12mm}}
    c@{\hspace{8mm}}
    c@{\hspace{4mm}}c@{}c@{\hspace{8mm}}
    c@{\hspace{4mm}}c@{}c@{\hspace{8mm}}
    c@{\hspace{4mm}}c
    @{}}
        & & \multicolumn{2}{c}{ACL} & & \multicolumn{2}{c}{GENIA} & & \multicolumn{2}{c}{ScienceIE} \\
        \cmidrule{3-4}\cmidrule{6-7}\cmidrule{9-10}
        Model (\#Params)& Fine-tuned on & exact $F_1$ & partial $F_1$ & & exact $F_1$ & partial $F_1$ & & exact $F_1$ & partial $F_1$ \\
\toprule
\multirow{4}{=}{\centering BERT Base\\(110M)} & UA-SS & $\mathbf{58.22}$ & $\mathbf{77.36}$  & & $\mathbf{53.18}$ & $79.38$  & & $46.79$ & $66.59$ \\
\cdashlinelr{2-10}
 & ACL & $-$ & $-$ & &  $52.05$ & $\mathbf{82.49}$  & & $47.88$ & $69.97$ \\
 & GENIA & $45.97$ & $61.53$ & &  $-$ & $-$   & & $\mathbf{48.50}$ & $\mathbf{69.84}$ \\
 & ScienceIE & $38.28$ & $54.92$ & &  $46.91$ & $73.16$  & & $-$ & $-$ \\
\midrule
\multirow{4}{=}{\centering ELECTRA Small\\(14M)} & UA-SS & $\mathbf{58.00}$ & $\mathbf{77.41}$  & & $\mathbf{53.44}$ & $80.01$  & & $44.68$ & $65.58$ \\
\cdashlinelr{2-10}
 & ACL & $-$ & $-$  & &  $50.84$ & $\mathbf{81.33}$ & &  $44.21$ & $67.57$ \\
 & GENIA & $46.65$ & $67.21$ & &  $-$ & $-$   & & $\mathbf{45.79}$ & $\mathbf{68.83}$ \\
 & ScienceIE & $42.58$ & $66.02$ & &  $43.48$ & $76.77$  & & $-$ & $-$ \\
\midrule
\multirow{4}{=}{\centering ELECTRA XSmall\\(7M)} & UA-SS & $\mathbf{49.83}$ & $\mathbf{72.78}$  & & $\mathbf{45.35}$ & $\mathbf{74.83}$  & & $\mathbf{40.32}$ & $\mathbf{62.39}$ \\
\cdashlinelr{2-10}
 & ACL & $-$ & $-$   & & $31.13$  & $59.99$  & & $28.79$ & $58.20$ \\
 & GENIA & $29.81$ & $58.17$ & &  $-$ & $-$  &  & $30.00$ & $59.61$ \\
 & ScienceIE & $20.60$ & $33.53$  & & $39.95$ & $68.63$  & & $-$ & $-$ \\
    \end{tabular}
    \caption{\label{tab:result_summary_all}Results for the generalization of multiple transformer models
             that are fine-tuned on the \textbf{weakly annotated} dataset based on the Semantic Scholar corpus (annotated with UA, denoted as UA-SS) and evaluated on the ACL, GENIA, and ScienceIE datasets, respectively. Transformer models fine-tuned using our automatically generated dataset perform better than their counterparts fine-tuned using the other datasets.}
\end{table*} 

\subsection{Results}

In Table~\ref{tab:result_summary_train}, we first compare the performance
(expressed as exact and partial $F_1$ scores that count only exact or partial
matches as true positives) of our fully \textit{Unsupervised Annotator} to the
performance obtained by fine-tuning transformer-based models with the manual
annotations present in the original training sets. Without relying on any human
annotation, our UA delivers comparable or even better results than the ELECTRA
XSmall in ACL and ScienceIE, respectively. These results show that the UA
represents a very \textit{competitive baseline} for domains where annotations are
not available.

Further, we are interested in understanding whether transformer-based models fine-tuned
with human annotations can generalize across domains. We also evaluate if the
availability of weakly supervised labels generated by our \textit{Unsupervised
Annotator} over a large and broad corpus (i.e., Semantic Scholar) could lead to
models with higher generalization capabilities. In
Table~\ref{tab:result_summary_all} we report the exact and partial $F_1$ scores
for the ACL, GENIA, and ScienceIE datasets, and the transformer-based model
fine-tuned with the output of our \textit{Unsupervised Annotator}~(UA-SS). 
This setup simulates the
problem~of bootstrapping an annotator for a specific domain~for which in-domain human
labels are not available.
 


On the ACL corpus, the UA-SS-based model clearly outperforms the GENIA-based
and ScienceIE-based models.  On the GENIA corpus, the UA-SS-based model and
the ACL-based model perform equally well.  On the ScienceIE corpus, all
models perform equally with a slight tendency towards the GENIA-based
model.

Overall, it can be said that the UA-SS-based approach is a valid starting point
to bootstrap a system in a no-resource scenario.
Table~\ref{tab:result_summary_train} shows that the F1 score gap between models
trained with in-domain manually annotated data and the UA-SS-based approach is
lower for smaller models.

\begin{figure}[!t]
	\centering
    \includegraphics[width=0.49\textwidth,height=2.75cm]{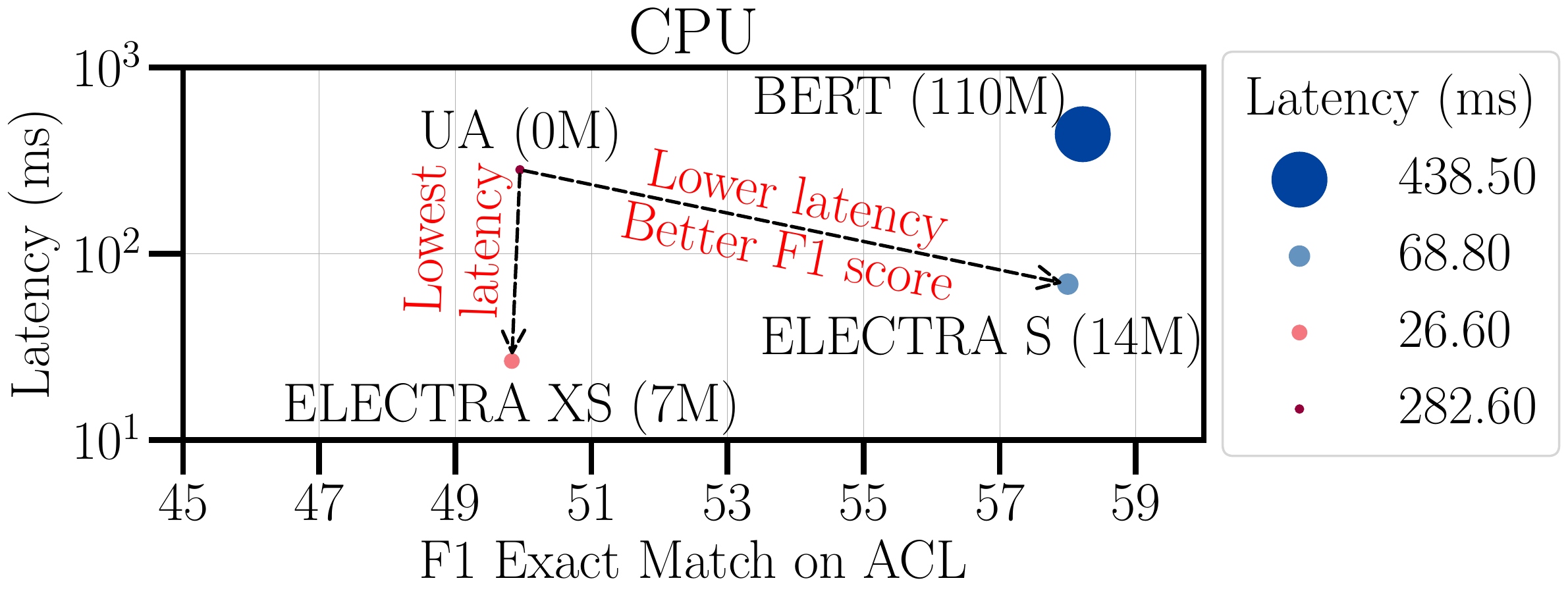}\\
	\includegraphics[width=0.49\textwidth,height=2.75cm]{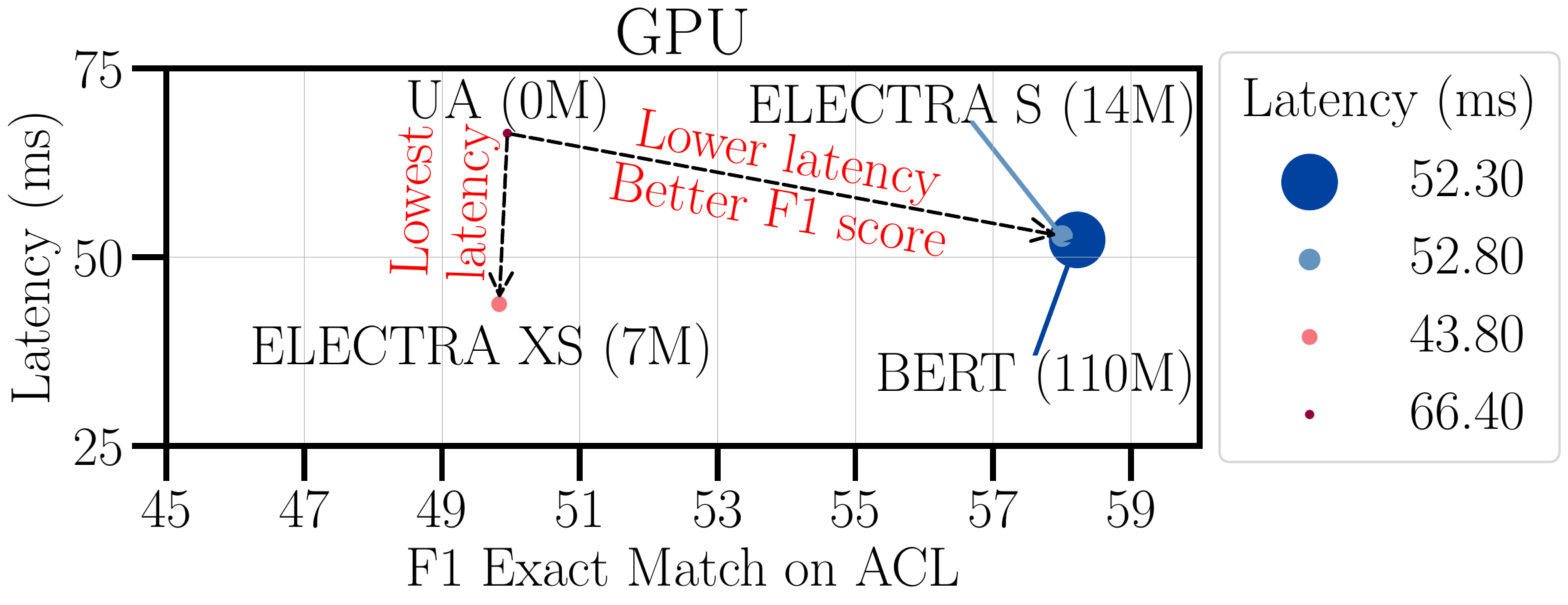}
	\caption{Average inference latency on CPU (top) and GPU (bottom) on the ACL dataset. We note in parenthesis the number of trainable parameters of the models. By fine-tuning over the output of the UA, we achieve~lower latency and higher F1 scores. The lowest inference latency, 26.6 ms, is achieved on CPU.}
    \label{fig:result_inference_latency}
\end{figure}

Now, we compare the \textit{Unsupervised Annotator} with the models fine-tuned with its output to evaluate our two-step approach in terms of F1 score \textit{and} inference latency. Figure~\ref{fig:result_inference_latency} reports the average inference latency
for models~(fine-tuned with the UA-SS training data) over sentences from the
ACL dataset with a batch of size 1 using a NVIDIA Tesla V100 and a single core
of a Xeon E5-2690 v4 (similar trends on the other datasets). 
While the inference latency has similar orders of magnitude across models with GPU acceleration, the minimum inference time of 26.6 ms can be
obtained on a \textit{single CPU core} using the ELECTRA XSmall model.
Therefore, our approach is particularly attractive in all cases where
inference accelerators~(e.g., GPUs) are not available. Additionally, the results
highlight that by fine-tuning over the output of the UA, the latency
can be \textit{reduced} by \textit{4 to 10 times}, while providing comparable
or even better F1 scores.
Having the option to generate a large amount of training data for fine-tuning is
an extremely useful property that enables the creation of very small models
offering low inference times even without using GPU acceleration.






\subsection{Lessons learned}

In this work, we have demonstrated that, while the value of in-domain labels is
without any doubt the best way to increase predictive quality, fully
unsupervised approaches are often the only viable option to bootstrap a term
extractor that has to generalize across very diverse domains. Additionally,
while the practicality of ML solutions is often underestimated, we have shown
that having a modular system can not only provide greater flexibility in deployments,
but can also allow to boost time predictive performance and inference~latency at the same.


\section{Conclusion}

In this paper, we described an effective term extraction approach that uses a
\textit{fully unsupervised annotator} to generate training data to fine-tune
transformer models. This approach reduces the inference time of the
unsupervised annotator, without decreasing its performance, and allows the
flexibility to pick the right trade-off between latency and F1 score. The
latency-optimized models are less than 30 Megabytes in size, provide inference
latencies lower then 30 ms even \textit{without} GPUs, while exhibiting a competitive F1 score compared to the models fine-tuned with manually annotated
data.

\bibliographystyle{acl_natbib}



\end{document}

%% file: tikz.tex
\usepackage{tikz}

\usepackage{pgfplots}
\pgfplotsset{compat=1.16}

\usetikzlibrary{
  arrows.meta, 
  fit, 
  positioning, 
  calc,
  shapes
}

\tikzset{
  amodule/.style={
    rectangle,draw,thick,
    text depth=.25ex,
    text centered,
    minimum width=2cm,
    minimum height=1cm
  },
  sarr/.style={ 
    -{Straight Barb[angle=60:2pt 3]}, 
    thick, 
  }
}